\documentclass{article}

\usepackage{microtype}
\usepackage{graphicx}
\usepackage{subcaption}
\usepackage{booktabs} 

\usepackage{hyperref}
\usepackage{multirow}

\usepackage[preprint]{icml2026}

\usepackage{amsmath}
\usepackage{amssymb}
\usepackage{mathtools}
\usepackage{amsthm}

\usepackage[capitalize,noabbrev]{cleveref}

\theoremstyle{plain}

\theoremstyle{definition}

\theoremstyle{remark}

\usepackage[textsize=tiny]{todonotes}

\definecolor{purple}{RGB}{128, 0, 128}
\definecolor{LightRed}{rgb}{1,0.92,0.92}
\definecolor{LightOrange}{rgb}{1,0.95,0.88}
\definecolor{LightYellow}{rgb}{1.0,1.0,0.84}
\definecolor{LightGreen}{rgb}{0.9,1.0,0.88}
\definecolor{LightCyan}{rgb}{0.9,1,1}
\definecolor{LightBlue}{rgb}{0.9,0.94,1}
\definecolor{LightIndigo}{rgb}{0.92,0.9,1}
\definecolor{LightMagenta}{rgb}{0.96,0.86,1}
\definecolor{DirtyWhite}{rgb}{0.96,0.96,0.96}

\DeclareSymbolFont{extraup}{U}{zavm}{m}{n}
\DeclareMathSymbol{\varheart}{\mathalpha}{extraup}{86}
\DeclareMathSymbol{\vardiamond}{\mathalpha}{extraup}{87}
\DeclareMathSymbol{\varclubsuit}{\mathalpha}{extraup}{88}

\icmltitlerunning{MoE Pathfinder: Trajectory-driven Expert Pruning}

\begin{document}

\twocolumn[
  \icmltitle{MoE Pathfinder: Trajectory-driven Expert Pruning}





    \begin{icmlauthorlist}
    Xican Yang$^{{\spadesuit}}$, \hspace{0.1cm}
    Yuanhe Tian$^{\varheart}$, \hspace{0.1cm}
    Yan Song$^{{\spadesuit}}$ \\
    \vspace{0.2cm}
    $^{\spadesuit}$University of Science and Technology of China \hspace{0.1cm}
    $^{\varheart}$Zhongguancun Institute of Artificial Intelligence \\
    \vspace{0.2cm}
    $^{\spadesuit}$\texttt{yxc15759879600@mail.ustc.edu.cn} \hspace{0.1cm}
    $^{\varheart}$\texttt{tianyuanhe@zgci.ac.cn} \hspace{0.1cm}
    $^{\spadesuit}$\texttt{clksong@gmail.com}
    \vspace{0.2cm}
    \end{icmlauthorlist}




]



\printAffiliationsAndNotice{}  

\begin{abstract}

Mixture-of-experts (MoE) architectures used in large language models (LLMs) achieve state-of-the-art performance across diverse tasks yet face practical challenges such as deployment complexity and low activation efficiency. Expert pruning has thus emerged as a promising solution to reduce computational overhead and simplify the deployment of MoE models. However, existing expert pruning approaches conventionally rely on local importance metrics and often apply uniform layer-wise pruning, leveraging only partial evaluation signals and overlooking the heterogeneous contributions of experts across layers. To address these limitations, we propose an expert pruning approach based on the trajectory of activated experts across layers, which treats MoE as a weighted computation graph and casts expert selection as a global optimal path planning problem. Within this framework, we integrate complementary importance signals from reconstruction error, routing probabilities, and activation strength at the trajectory level, which naturally yields non-uniform expert retention across layers. Experiments show that our approach achieves superior pruning performance on nearly all tasks compared with most existing approaches.

\end{abstract}

\renewcommand{\thefootnote}{\arabic{footnote}}


\section{Introduction}

Large language models (LLMs) almost dominate all tasks across a wide range of scenarios with consistently strong performance \cite{brown2020language,wei2022emergent,liu2023visual,li2023starcoder,lin2023evolutionary,touvron2023llama,achiam2023gpt}.
The recent progress on LLMs shows that the mixture-of-experts (MoE) architecture has emerged as a key paradigm because of its superior performance-compute trade-off \cite{fedus2022switch,jiang2024mixtral,dai2024deepseekmoe,liu2024deepseek,meta2025llama,team2025kimi,zeng2025glm},
where a sparse activation mechanism is utilized with a small subset of experts activated for each input, enabling large parameter scales without proportional computational overhead \cite{lepikhin2020gshard,artetxe2021efficient}.
Owing to its computational sparsity, MoE models require loading all expert weights during inference, no matter whether an expert is activated, resulting in a substantial random access memory (RAM) and storage burden\footnote{E.g., Mixtral-8x7B \cite{jiang2024mixtral} activates only a few experts per pass but must keep all parameters resident in RAM.}.
Furthermore, in task-specific or domain-adapted scenarios, only a small subset of experts is usually enough to achieve strong performance, while activating all experts yields marginal gains but greatly reduces inference efficiency \cite{su2025unveiling,lo2025closer}.
Therefore, to preserve the specialization of MoE models while improving inference efficiency, it is important to adopt expert pruning strategies that effectively operate on the MoE architecture.

\begin{figure*}[t]
  \centering
  \includegraphics[width=1\linewidth, trim=0 10 0 0]{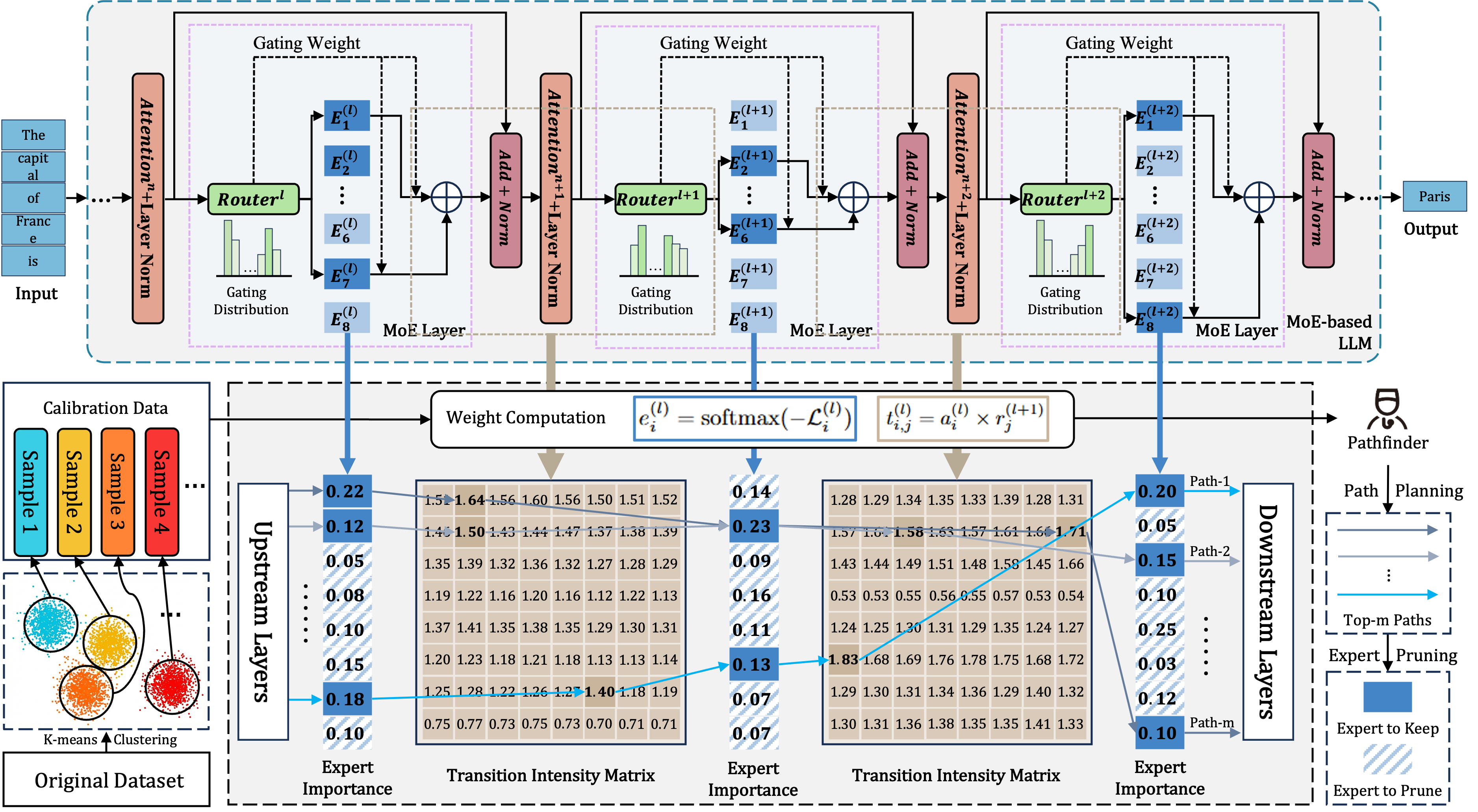}
  \caption{
    Overview of the proposed expert pruning framework based on the trajectory of activated experts across layers.
    The top part presents the standard multi-layer MoE architecture, where tokens are dynamically routed to sparse experts.
    The bottom part shows our pruning approach that reformulates the MoE as a directed weighted graph. 
    In our approach, we firstly compute transition intensities (edge weights) and expert importance scores (node weights) based on routing probabilities, expert activations, and reconstruction loss, respectively. 
    Then, we use a global path planning algorithm to identify the top-ranked optimal inference trajectories.
    Experts located on these critical paths are retained (highlighted in purple), while the remaining redundant experts are pruned.
    }
\label{fig:approach}
\vspace{-0.2cm}
\end{figure*}

Existing studies on compressing MoE architectures include dynamic expert selection, expert merging, and expert pruning.
Dynamic expert selection introduces zero-computation experts or estimates token-level importance to dynamically adjust the number of activated experts during inference, but it does not reduce parameter counts \cite{jin2024moe++,raposo2024mixture,zeng2024adamoe,yue2024ada,team2025longcat}.
Expert merging reduces the number of experts by linearly combining weights, but it often harms expert specialization \cite{li2023merge,he2023merging,chen2024retraining,li2025sub}.
Expert pruning removes some experts and becomes the mainstream direction because it lowers memory and computational costs while preserving expert specialization \cite{guo2025cluster,zhang2025mone}.
Existing expert pruning approaches are broadly categorized into activation-based strategies and perturbation-based strategies.
Activation-based strategies typically estimate expert importance using routing probabilities and expert activations \cite{muzio2024seer,dong2025domain,lasby2025reap}.
While computationally efficient, these approaches rely on local statistics and fail to capture the experts’ contribution to global task performance.
In contrast, perturbation-based approaches assess importance by measuring expert output norms or performance discrepancies when experts are removed \cite{cao2024condense,lu2024not}.
These approaches offer a more comprehensive view of expert influence but require multiple forward passes or reconstruction-based sensitivity analysis, which incurs high computational costs.
Moreover, most existing approaches \cite{zhang2025mone,lee2025stun} apply a uniform pruning ratio across all layers, ignoring the heterogeneity of expert importance among layers,
leading to insufficient resources for critical layers and insufficient compression for redundant ones, ultimately degrading overall performance and efficiency.
Therefore, existing expert pruning approaches either depend on local heuristics or require expensive computations, making it difficult to achieve a globally consistent yet efficient evaluation of expert importance.
As a result, there is a pressing need for a pruning strategy that integrates multi-dimensional information and accounts for cross-layer heterogeneity in a computationally efficient way.
Intuitively, the routing mechanism in MoE inherently performs a path selection across layers, which naturally motivates reformulating expert selection as a global optimization problem on a graph.

In this paper, we propose an expert pruning approach that leverages the activation trajectories of experts across layers, which treats MoE as a weighted computation graph and casts expert selection as a global optimal path planning problem.
We firstly model the MoE architecture as a layered graph, where experts act as nodes and inter-layer interactions are captured through transition intensity matrices.
Within this graph, we quantify information flow using two complementary metrics: transition intensities and expert importance scores.
Leveraging these metrics, we cast expert selection as identifying the top globally significant computation paths through a dynamic-programming-based path exploration scheme.
By aggregating critical experts appearing in these high-value trajectories across calibration samples, we construct a unified pruning mask to realize structured MoE compression.
This graph-driven global evaluation framework integrates activation strength, routing behaviors, and reconstruction-based expert importance, enabling adaptive cross-layer non-uniform pruning while preserving the model’s specialized capabilities.
Experiment results show that our approach substantially reduces inference cost and memory usage while maintaining or surpassing baseline performance on benchmark datasets.

\section{The Approach}
\label{sec:approach}

We propose an expert pruning approach based on the trajectory of activated experts across layers that performs expert-level pruning for task-specific MoE models.
The complete workflow is illustrated in Figure \ref{fig:approach}, where the upper part outlines the standard MoE architecture and the lower part visualizes how the proposed trajectory-driven approach is constructed and used for pruning via global path planning.
In contrast to conventional pruning schemes that assess experts independently and often in a layer-wise manner, our approach reformulates the MoE computation as a structured graph. 
On this graph, we compute data-driven weights that capture both the intrinsic contribution of each expert and the cross-layer information flow between experts. 
We then perform global path planning to identify the paths most critical to the target task and prune away experts that do not participate in any of these high-importance paths.
In the following sections, we present the details of the graph construction, the computation of trajectory weights, and the global-path-planning strategy for expert pruning.

\subsection{Graph Construction}
\label{sec:graph_construction}

We formalize the MoE architecture as a layer-wise graph that captures all feasible routes of information flow during inference. The purpose of this representation is to make expert-level dependencies across layers visible so that pruning decisions are made at the global path level rather than in an isolated, per-layer manner.
Formally, the MoE model contains $L$ expert layers. At layer $l$, we denote its $N_e$ experts by the set ${E^{(l)}_1, \dots, E^{(l)}_{N_e}}$. Each expert corresponds to a node in the trajectory.
Stacking all layers together yields a directed acyclic graph composed of $L$ groups of expert nodes, where edges only appear between adjacent layers.
A complete inference trajectory is thus visualized as selecting one expert per layer, connected through the transition intensities across successive layers.
This graph abstraction allows the information-propagation pattern to be explicit and establishes a unified foundation for the subsequent weight computation and global path–based expert pruning.

\subsection{Weight Computation}
\label{sec:weight}
Once the graph is constructed, we assign two weight metrics: the transition intensity between adjacent experts and the importance score of each expert.
Transition intensity models the strength of cross-layer information flow, while expert importance reflects each expert’s individual contribution.
Together, these metrics define the quality of computation paths and allow pruning to consider both local expert quality and global path structure.
All weights are computed using sample-level statistics from a calibration dataset.

Before describing the construction of these weights, we introduce notations for clarity.
At layer $l$, the MoE module contains $N_e$ experts with weight matrices $\mathbf{W}^{(l)}_1, \dots, \mathbf{W}^{(l)}_{N_e}$.
Each expert weight matrix has a shape of $(d_{\text{out}}, d_{\text{in}})$, where $d_{\text{in}}$ denotes the input dimensionality of experts and $d_{\text{out}}$ denotes the output dimensionality.
The router associated with layer $l$ is parameterized by a matrix $\mathbf{R}^{(l)} \in \mathbb{R}^{N_e \times d_{\text{in}}}$, which maps the layer input representation into routing logits over the $N_e$ experts.
Given an input sample $X$ from the calibration dataset $D$, we denote the hidden state entering layer $l$ as $\mathbf{H}^{(l-1)} \in \mathbb{R}^{N_x \times d_{\text{in}}}$, where $N_x$ is the number of tokens in the sample.
For the first layer, we have $\mathbf{H}^{(0)} = \mathbf{X}$.
In the following text, we detail how the transition intensities and expert importance scores are computed.

\paragraph{Transition Intensity}
To encode the potential flow of computations across layers, we introduce a fully connected transition intensity matrix $\mathbf{T}^{(l)} = \left(t^{(l)}_{i,j}\right)_{N_e \times N_e}$ between layer $l$ and layer $l+1$, where each entry $t^{(l)}_{i,j}$ represents the transition strength from expert $E^{(l)}_i$ to expert $E^{(l+1)}_j$.
To make this quantity more interpretable, we decompose the transition into two components, namely, the upstream activation strength $a^{(l)}_i$ and the downstream routing probability $r^{(l+1)}_j$.

Specifically, for the upstream activation strength, it quantifies how strongly expert $i$ contributes to token representations at layer $l$.
For the $k$-th token, the output magnitude of expert $i$ is computed as
\begin{equation}
a^{(l)}_{i,k} = \left\| \mathbf{H}^{(l-1)}_k (\mathbf{W}^{(l)}_i)^\top \right\|_2
\end{equation}
where $\left\| \cdot \right\|_2$ computes the $\ell_2$-norm of a vector.
Averaging over all tokens yields the upstream activation strength:
\begin{equation}
a^{(l)}_i = \frac{1}{N_x} \sum_{k=1}^{N_x} a^{(l)}_{i,k}
\end{equation}
This term captures the amount of signal emitted by expert $i$.

For downstream routing probability, we measure how likely expert $j$ is to be selected by the router in layer $l+1$.
For each token, its routing probability $r^{(l+1)}_{j,k}$ is computed as
\begin{equation}
r^{(l+1)}_{j,k} =
\left[\mathrm{softmax}\left(\mathbf{H}^{(l)}_k (\mathbf{R}^{(l+1)})^\top\right)\right]_j
\end{equation}
Averaging over tokens gives the downstream routing preference:
\begin{equation}
r^{(l+1)}_j = \frac{1}{N_x} \sum_{k=1}^{N_x} r^{(l+1)}_{j,k}
\end{equation}
This term captures the likelihood that expert $j$ will receive information in the next layer.

Combining the upstream activation strength and downstream routing probability, we obtain the overall transition intensity:
\begin{equation} \label{eq:transition-intensity}
t^{(l)}_{i,j} = a^{(l)}_i \times r^{(l+1)}_j
\end{equation}
This factorization highlights that a strong transition occurs only when the upstream expert emits a large activation and the downstream expert has a high routing probability.

\paragraph{Expert Importance}

For expert $i$ in layer $l$, we define an expert importance score $e^{(l)}_i$ that quantifies its individual contribution to the layer’s output.
To obtain this score, we measure how well expert $i$ alone is able to reconstruct the original output of layer $l$.

Firstly, we denote the original output of layer $l$ for token $k$ as $\mathbf{Y}^{(l)}_k$.
When only expert $i$ is active, its reconstructed output is
\begin{equation}
\hat{\mathbf{Y}}^{(l)}_{k,i} = \mathbf{H}^{(l-1)}_k (\mathbf{W}^{(l)}_i)^\top
\end{equation}
This provides a token-wise view of how much expert $i$ alone contributes to the layer computation.
Next, the discrepancy between the original output and the expert-only reconstruction is quantified by the per-expert reconstruction loss:
\begin{equation}
\mathcal{L}^{(l)}_i
= \frac{1}{N_x} \sum_{k=1}^{N_x}
\left\| \mathbf{Y}^{(l)}_k - \hat{\mathbf{Y}}^{(l)}_{k,i} \right\|_2^2
\end{equation}
A smaller loss indicates that expert $i$ more faithfully captures the behavior of the full MoE layer.
Then, we convert reconstruction losses into importance scores via a softmax transformation:
\begin{equation}
e^{(l)}_i = \mathrm{softmax}(-\mathcal{L}^{(l)}_i)
\end{equation}
assigning greater importance to experts who more closely approximate the original output.

Different from intermediate layers, the first and last layers lack upstream or downstream routing signals, respectively.
To capture these missing directional flows, we incorporate additional correction terms.
For the first layer:
\begin{equation}
e^{(1)}_i
= \mathrm{softmax}(-\mathcal{L}^{(1)}_i)\times
\frac{1}{N_x} \sum_{k=1}^{N_x}
\left[
\mathrm{softmax}\big( \mathbf{X}_k (\mathbf{R}^{(1)})^\top \big)
\right]_i
\end{equation}
where the notation $[\cdot]_i$ extracts the $i$-th entry of the softmax vector, indicating the routing probability that the router of the first layer assigns to expert $i$.
For the last layer:
\begin{equation}
e^{(L)}_i
= \mathrm{softmax}(-\mathcal{L}^{(L)}_i)\times
\frac{1}{N_x} \sum_{k=1}^{N_x}
\left\| \mathbf{H}^{(L-1)}_k (\mathbf{W}^{(L)}_i)^\top \right\|_2
\end{equation}
which incorporates the final activation strength produced by expert $i$.
These adjustments ensure that the importance scores for boundary layers reflect the same directional flow semantics as those in intermediate layers.

\subsection{Pruning via Global Path Planning}
\label{sec:pruning}

After constructing the weighted graph, we identify essential experts by analyzing the global computation paths that span from the first MoE layer to the last.
The key idea is that each feasible path represents a potential sequence of expert activations during inference.
We therefore search for the top-$m$ most important paths and retain only the experts that appear along these paths.
To enable this selection, we assign a weight to each feasible path based on the expert- and transition-level quantities obtained from Sec. \ref{sec:weight}.
Specifically, each path weight integrates two complementary components: the transition intensities $t^{(l)}_{i,j}$, which capture the strength of information flow from layer $l$ to $l{+}1$, and the expert importance scores $e^{(l)}_i$, which reflect the contribution of each expert within its layer.
Together, these quantities provide a unified assessment of how effectively a path propagates information through the network.

Specifically, let $\mathcal{P}$ denote the set of all feasible paths that select one expert per layer.
For any path $p \in \mathcal{P}$, its weight $w_p$ is defined as $w_p = \prod_{l=1}^{L-1} t^{(l)}_{\,i_l,i_{l+1}} \prod_{l=1}^{L} e^{(l)}_{\,i_l}$,
where $i_l$ denotes the expert selected by path $p$ at layer $l$.
This multiplicative structure reflects that a path is important only when it consistently involves influential experts and strong transitions between layers.
To avoid numerical instability caused by multiplying many small values, the path weight is evaluated in the logarithmic domain, yielding the following additive formulation:
\begin{equation}
\label{eq:log-path-weight}
\log w_p
=
\sum_{l=1}^{L-1} \log t^{(l)}_{\,i_l,i_{l+1}}
+
\sum_{l=1}^{L} \log e^{(l)}_{\,i_l}
\end{equation}
The top-$m$ path selection problem for MoE pruning is equivalent to identifying the $m$ paths with the highest log-weights.
Since every feasible path selects exactly one expert per layer, all paths have identical length.
This ensures that the multiplicative weighting does not inherently favor shorter paths, and differences in path weights arise solely from the quality of the experts and transitions along the path.

Since enumerating all feasible paths is computationally infeasible, we adopt a dynamic programming procedure that incrementally constructs the highest-weight partial paths.
We define a prefix path as a partial path spanning layers $1$ through $l$, represented by the sequence of expert indices selected up to layer $l$.
For each expert node $v$ in layer $l$, we maintain a priority queue $Q_v$ containing up to $m$ prefix paths that terminate at $v$, each stored along with its log-weight.
Then, we process nodes layer by layer in topological order.
Specifically, for every incoming edge from expert $i_l$ in layer $l$ to expert $i_{l+1}$ in layer $l{+}1$, each prefix path in $Q_{i_l}$ is extended by appending expert $i_{l+1}$.
The log-weight of the extended prefix path is computed as
\begin{equation}
\label{eq:extended-path-weight}
\log w_{\text{new}}
=
\log w_{\text{old}}
+
\log t^{(l)}_{\,i_l,i_{l+1}}
+
\log e^{(l+1)}_{\,i_{l+1}}
\end{equation}
where $\log w_{\text{old}}$ denotes the log-weight of the prefix path before extension.
For each node, only the $m$ extended paths with the highest weights are retained in its priority queue.

This dynamic programming procedure is applied independently to each sample in the calibration dataset, producing a set of top-$m$ full paths per sample.
For a given sample $X$, let $\mathcal{P}_{m,X}$ denote its selected top-$m$ paths.
The experts preserved for the sample $X$ are defined as the union of all experts appearing in these paths $\mathcal{E}_{\text{keep},X}=\bigcup_{p \in \mathcal{P}_{m,X}}\mathcal{V}(p)$.
At the dataset level, the final set of experts retained by the pruning procedure is the union of per-sample expert sets $\mathcal{E}_{\text{keep}}=\bigcup_{X\in\mathcal{D}}
\mathcal{E}_{\text{keep},X}$.
To apply pruning to the MoE model, a binary mask is constructed for each layer to disable experts not included in $\mathcal{E}_{\text{keep}}$:
\begin{equation}
\label{eq:binary-mask}
M^l_i =
\begin{cases}
1, & E^{(l)}_i \in \mathcal{E}_{\text{keep}} \\
0, & \text{otherwise}
\end{cases}
\end{equation}
After determining the mask, the model is pruned by removing the parameters of unselected experts and by retaining only the rows in the routing matrix that correspond to surviving experts.
This pruning procedure significantly reduces both the memory footprint and computational cost during inference while preserving the most critical expert pathways.

\section{Experiment Settings}

\subsection{Datasets}
\label{sec:dataset}

To comprehensively evaluate our pruning approach, we conduct experiments on six widely used benchmark datasets that target different aspects of model performance.
These datasets include MMLU \cite{hendrycks2020measuring}, HellaSwag \cite{zellers2019hellaswag}, WinoGrande \cite{sakaguchi2021winogrande}, {ARC} \cite{clark2018think}, {GSM8K} \cite{cobbe2021gsm8k} and MedQA \cite{jin2021disease} that cover both general and domain-specific fields.\footnote{We obtain MMLU, HellaSwag, WinoGrande, ARC, GSM8K and MedQA from \url{https://huggingface.co/datasets/cais/mmlu}, \url{https://huggingface.co/datasets/Rowan/hellaswag}, \url{https://huggingface.co/datasets/allenai/winogrande}, \url{https://huggingface.co/datasets/allenai/ai2_arc}, \url{https://huggingface.co/datasets/openai/gsm8k} and \url{https://huggingface.co/datasets/GBaker/MedQA-USMLE-4-options}.}
Specifically, MMLU is a multiple-choice benchmark covering 57 subjects (e.g., humanities, STEM), testing both factual knowledge and reasoning ability.
HellaSwag is a commonsense reasoning benchmark where models select the most plausible sentence ending from multiple choices, known for being easy for humans but challenging for machines.
WinoGrande is a large-scale dataset for assessing commonsense reasoning through pronoun resolution tasks, designed to minimize biases via debiasing techniques.
GSM8K is a dataset of high-quality, linguistically diverse grade-school math word problems.
MedQA is a medical question-answering (QA) dataset derived from United States Medical Licensing Examination questions in a four-choice format, covering clinical diagnosis and treatment scenarios in the medical domain.
ARC contains grade-school science questions in multiple-choice format, requiring qualitative reasoning across physics, biology, and chemistry.
The number of instances in the training, validation, and test datasets is reported in Table \ref{tab:data}.

\begin{table}[t]
\centering
\caption{
The number of instances in the training, validation, and test sets of all datasets used in our experiments.
}
\label{tab:data}
\vspace{-0.2cm}
\begin{tabular}{l|rrr} 
\toprule
\textbf{Dataset} & \textbf{Training} & \textbf{Validation} & \textbf{Test}\\
\midrule
MMLU
& 99,842
& 1,531
& 14,042\\
HellaSwag
& 39,905
& 10,042
& 10,003 \\
WinoGrande
& 40,398
& 1,267
& 1,767 \\
GSM8K
& 7,473
& -
& 1,319 \\
MedQA
& 10,178
& -
& 1,273 \\
ARC
& 1,119
& 299
& 1,172 \\
\bottomrule
\end{tabular}
\vspace{-0.5cm}
\end{table}

\begin{table*}[t]
\centering
\caption{
Overall performance comparison of our expert pruning approach against baselines.
This table compares the performance of ``Full'' (unpruned), ``Random'' (random baseline), and ``Ours'' (our approach) for Mixtral-8x7B and Mixtral-8x7B-Instruct models.
All pruning approaches are set to 50\% expert sparsity.
All reported metrics are accuracy scores, where higher is better.
``Avg.'' refers to the average accuracy across MMLU, HellaSwag, WinoGrande, ARC, GSM8K, and MedQA.
}
\label{tab:usage}
\vspace{-0.2cm}
\begin{tabular}{l|ccc|ccc|c}
\multicolumn{8}{c}{(a) Mixtral-8x7B} \\
\toprule
\textbf{Approach}
& \textbf{MMLU}
& \textbf{HellaSwag}
& \textbf{WinoGrande}
& \textbf{ARC}
& \textbf{GSM8K}
& \textbf{MedQA}
& \textbf{Avg.} \\
\midrule
Full
& 67.88
& 64.83
& 76.08
& 56.48
& 58.45
& 62.37
& 64.35 \\
Random
& 25.39
& 40.62
& 56.98
& 23.03
& 1.13
& 26.08
& 28.87 \\
Ours
& 54.31
& 56.51
& 74.19
& 49.66
& 38.67
& 49.25
& 53.77 \\
\bottomrule
\multicolumn{8}{c}{(b) Mixtral-8x7B-Instruct} \\
\toprule
\textbf{Approach}
& \textbf{MMLU}
& \textbf{HellaSwag}
& \textbf{WinoGrande}
& \textbf{ARC}
& \textbf{GSM8K}
& \textbf{MedQA}
& \textbf{Avg.} \\
\midrule
Full
& 68.81
& 67.61
& 76.37
& 62.54
& 65.04
& 60.88
& 66.88 \\
Random
& 31.12
& 46.76
& 56.83
& 29.61
& 1.90
& 30.87
& 34.52 \\
Ours
& 57.66
& 60.42
& 73.01
& 53.18
& 30.10
& 46.74
& 53.52 \\
\bottomrule
\end{tabular}
\vspace{-0.3cm}
\end{table*}

For the construction of the calibration set, we design and construct distinct calibration datasets tailored to tasks from various domains.
Specifically, the calibration data for each task is derived from the training set of its corresponding dataset.
To ensure sufficient diversity in the calibration data, we apply k-means clustering to each dataset independently, partitioning it into $K$ clusters (the specific value of $K$ is provided in Section \ref{sec:implementation-detail}).
From each cluster, we select the sample closest to the cluster centroid to construct the representative calibration data.
Furthermore, following the approach proposed by \citet{dong2025domain}, for each sample in the obtained calibration data, we replace its desired sample output by the output of an LLM with the sample input, which is demonstrated to be effective in LLM pruning.

\subsection{Implementation Details}
\label{sec:implementation-detail}

We conduct experiments on two widely adopted MoE-based LLMs, namely, Mixtral-8x7B and Mixtral-8x7B-Instruct \cite{jiang2024mixtral}.\footnote{The LLMs are obtained from their official HuggingFace repositories \url{https://huggingface.co/}.}
Mixtral-8x7B is the general pre-trained base model designed for broad language understanding, while Mixtral-8x7B-Instruct is its instruction-tuned variant optimized for responding to user prompts and following instructions.
Both Mixtral-8x7B and Mixtral-8x7B-Instruct consist of 32 Transformer layers with a hidden dimension of 4,096, and each layer incorporates an MoE block containing eight experts from which two are activated per token.
For evaluation, we use the EleutherAI LM Harness\footnote{\url{https://github.com/EleutherAI/lm-evaluation-harness}} to assess the accuracy of different models.

For different tasks, we use different cluster numbers $K$ to partition the calibration dataset into clusters.
Specifically, we set $K\!=\!12$ for MMLU and WinoGrande, $K\!=\!20$ for HellaSwag, $K\!=\!10$ for ARC, $K\!=\!9$ for GSM8K, and $K\!=\!5$ for MedQA.\footnote{We try different values of $K$ and use the one that achieves the best performance for each dataset.}
To precisely control the expert coverage across varying models and compression ratios, which is crucial for subsequent pruning, we regulate the path planning process via a top-$m$ parameter.
For the Mixtral series, we set $m\!=\!1$ for pruning with an expert sparsity of 50\%, and increase it to 500 when performing pruning under a 25\% sparsity level.

\section{Results and Analysis}

\subsection{Overall Results}

\begin{table*}[t]
\centering
\caption{
Performance comparison of our pruning approach against existing MoE compression approaches for the Mixtral-8x7B model, with all approaches evaluated at 50\% expert sparsity.
``$^*$'' indicates that the data was obtained through replication, and was not taken directly from the original paper.
}
\label{tab:exsiting-approach}
\vspace{-0.2cm}
\begin{tabular}{ll|ccc|ccc|c} 
\toprule
\textbf{Technique}
&\textbf{Approach}
& \textbf{MMLU}
& \textbf{HellaSwag}
& \textbf{WinoGrande}
& \textbf{ARC}
& \textbf{GSM8K}
& \textbf{MedQA}
& \textbf{Avg.}    \\
\midrule
\multirow{3}{*}{Merging} &
\citet{li2023merge}$^*$
& 23.00
& 26.74
& 51.14
& 19.62
& 0.00
& 28.12
& 24.77 \\
&\citet{chen2024retraining}
& 48.95
& \textbf{57.81}
& 72.06
& 46.42
& -
& -
& - \\
&\citet{li2025sub}
& 48.00
& 57.00
& 72.00
& 45.00
& -
& -
& - \\
\midrule
\multirow{8}{*}{Pruning} &
\citet{he2024demystifying} 
& 25.54
& 42.50
& 56.99
& 22.35
& -
& - 
& - \\
&\citet{lu2024not} 
& 47.30
& 57.66
& 72.85
& 48.89
& -
& - 
& - \\
&\citet{muzio2024seer}
& 49.64
& -
& -
& -
& -
& -
& - \\
&\citet{lee2025stun}$^*$
& 24.19
& 25.83
& 50.51
& 23.12
& 0.00
& 24.83
& 24.75 \\
&\citet{jaiswal2025finding}$^*$
& 23.17
& 26.31
& 50.67
& 21.33
& 0.00
& 26.87
& 24.73 \\
&\citet{zhou2025dropping}$^*$
& 41.30
& 52.61
& 72.14
& 40.61
& 11.22
& 29.85
& 41.29 \\
&\citet{dong2025domain}$^*$
& 53.75
& 56.19
& 71.67
& 44.88
& 36.69
& 41.01
& 51.70 \\
\cmidrule{2-9}
& \textbf{Ours}
& \textbf{54.31}
& 56.51
& \textbf{74.19}
& \textbf{49.66}
& \textbf{38.67}
& \textbf{49.25}
& \textbf{53.77} \\
\bottomrule
\end{tabular}
\vspace{-0.3cm}
\end{table*}

\begin{table*}[t]
\centering
\caption{
Performance of our pruning approach under different expert sparsity levels (25\% and 50\%).
The performance is evaluated for Mixtral-8x7B and Mixtral-8x7B-Instruct.
}
\label{tab:expert-sparsity}
\vspace{-0.2cm}
\begin{tabular}{l|ccc|ccc|c}
\multicolumn{8}{c}{(a) Mixtral-8x7B} \\
\toprule
\textbf{Sparsity}
& \textbf{MMLU}
& \textbf{HellaSwag}
& \textbf{WinoGrande}
& \textbf{ARC}
& \textbf{GSM8K}
& \textbf{MedQA}
& \textbf{Avg.} \\
\midrule
25\%
& 59.85
& 62.14
& 75.62
& 52.56
& 44.20
& 60.80
& 59.20 \\
50\%
& 54.31
& 56.51
& 74.19
& 49.66
& 38.67
& 49.25
& 53.77 \\
\bottomrule
\multicolumn{8}{c}{(b) Mixtral-8x7B-Instruct} \\
\toprule
\textbf{Sparsity}
& \textbf{MMLU}
& \textbf{HellaSwag}
& \textbf{WinoGrande}
& \textbf{ARC}
& \textbf{GSM8K}
& \textbf{MedQA}
& \textbf{Avg.} \\
\midrule
25\%
& 63.13
& 65.17
& 75.22
& 59.04
& 54.59
& 59.31
& 62.74 \\
50\%
& 57.66
& 60.42
& 73.01
& 53.18
& 30.10
& 46.74
& 53.52 \\
\bottomrule
\end{tabular}
\vspace{-0.3cm}
\end{table*}

\begin{table*}[t]
\centering
\caption{
Ablation study on the information integration modalities, conducted for the Mixtral-8x7B model.
We analyze the impact of two weight metrics: ``TI'' (transition intensity) and ``IS'' (expert importance score).
"$\checkmark$" indicates the inclusion of the information modality, while "$\times$" indicates its exclusion.
}
\label{tab:ablation}
\vspace{-0.2cm}
\begin{tabular}{cc|ccc|ccc|c} 
\toprule
\textbf{IS}
& \textbf{TI}
& \textbf{MMLU}
& \textbf{HellaSwag}
& \textbf{WinoGrande}
& \textbf{ARC}
& \textbf{GSM8K}
& \textbf{MedQA}
& \textbf{Avg.}   \\
\midrule
\checkmark 
& \checkmark 
& 54.31
& \textbf{56.51}
& \textbf{74.19}
& \textbf{49.66}
& \textbf{38.67}
& \textbf{49.25}
& \textbf{53.77} \\
$\times$
&\checkmark 
& 54.01
& 53.90
& 72.85
& 46.07
& 38.06
& 46.98
& 51.98 \\
\checkmark
& $\times$
& \textbf{58.77}
& 54.09
& 66.85
& 41.55
& 0.70
& 38.18
& 43.36 \\
\bottomrule
\end{tabular}
\vspace{-0.3cm}
\end{table*}

We compare our proposed expert pruning approach against the original full model and random expert pruning baseline across two distinct LLMs (i.e., Mixtral-8x7B and Mixtral-8x7B-Instruct).
We summarize the main results in Table \ref{tab:usage}, where several observations are drawn from the results.
First, our approach outperforms the random pruning baseline across all models.
Furthermore, it approaches the performance of the full model on certain tasks, such as WinoGrande.
Even on the strictly evaluated GSM8K benchmark, our approach maintains a degree of performance preservation.
This phenomenon strongly proves that our approach is not merely removing experts, but possesses the ability to accurately identify and protect the model's core knowledge and routing logic.
Second, our approach is not only effective on the base model (Mixtral-8x7B) but also demonstrates superior performance compared to the random baseline on the instruction-tuned model (Mixtral-8x7B-Instruct), where expert specialization is more pronounced.
This proves that our approach is able to capture the critical contribution of experts to the model's overall representation learning, thus making it widely applicable to MoE model compression tasks across various training stages.

To further demonstrate the effectiveness of our approach, we compare the performance of our best model against existing MoE compression approaches for the Mixtral-8x7B model, with all approaches evaluated at $50\%$ expert sparsity. 
The results on various benchmark datasets are reported in Table \ref{tab:exsiting-approach}.
Our approach consistently outperforms existing MoE compression approaches across nearly all datasets and achieves the highest average accuracy (Avg.).
In particular, compared with our approach, \citet{lee2025stun} prune experts by solely examining the similarity of expert weights and router outputs, while they fail to account for the varying importance of experts across different tasks; \citet{lu2024not} exhaustively enumerates expert combinations to calculate the reconstruction error, which results in prohibitively high computational complexity; \citet{dong2025domain} utilizes router and output metrics for pruning but overlooks the reconstruction error of individual experts.
And all these approaches employ uniform pruning across all layers, which ignores the inherent heterogeneity in the importance distribution of the expert networks layer-by-layer.
Our approach shows its superiority to the aforementioned studies since, while maintaining a computational overhead close to a single forward pass, we integrate three distinct types of information—perturbation (reconstruction error), routing decisions, and activation strength—specifically tailored for the target task. 
Furthermore, we achieve layer-wise non-uniform expert pruning through a path-finding strategy.

\subsection{Effect of Expert Sparsity}

To systematically assess how the pruning ratio (i.e., expert sparsity) influences model performance, we extend our evaluation to multiple sparsity levels across different MoE architectures.
Specifically, we evaluate our approach under 25\% and 50\% expert retention on Mixtral-8x7B and Mixtral-8x7B-Instruct.
The complete results are reported here in Table \ref{tab:expert-sparsity}.
The following are some observations.
First, language understanding and knowledge-intensive tasks (e.g., MMLU, HellaSwag, MedQA) exhibit higher sensitivity to sparsity.
Both Mixtral-8x7B and Mixtral-8x7B-Instruct show clear performance reduction along increased sparsity.
An explanation is that these tasks depend on a broad set of experts, and higher pruning ratios may remove critical information pathways required for recalling distributed knowledge.
Second, for certain tasks, increasing retained experts from 50\% to 25\% yields only marginal performance gains.
For instance, in both Mixtral models, WinoGrande exhibit noticeably smaller performance improvements.
This phenomenon suggests that these tasks rely on a relatively compact subset of experts, with core information predominantly captured by a small number of specialists.
Consequently, even under a 50\% sparsity level, the experts retained through our path planning already cover the essential computation routes, and the additional specialists preserved at 25\% sparsity provide limited incremental benefit overall.

\subsection{Ablation Study}

To validate the effectiveness of our approach in information integration, we conduct an ablation study on the Mixtral-8x7B model to evaluate the impact of different information integration strategies used in the expert pruning approach.
Specifically, we analyze the impact of two distinct weight metrics derived in Sec. \ref{sec:weight}: the transition intensity (TI) and the expert importance score (IS).
To isolate the contribution of each weight, we evaluate variants where specific information is disregarded by setting the corresponding weight to a constant value of 1, thereby neutralizing its influence on the path planning and pruning process.
The complete results are presented in Table \ref{tab:ablation}.
Several observations are made from these results.
We observe that using only transition intensity provides a more balanced profile but fails to reach the peak average accuracy.
This indicates that transition intensity focuses on preserving the topological structure of expert connections, but may fail to retain some experts that store essential knowledge within the model.
Conversely, relying solely on the expert importance score leads to unstable performance; while it maintains high accuracy on a knowledge-heavy task (MMLU), it suffers a catastrophic failure in mathematical reasoning, significantly dragging down the average.
This catastrophic failure suggests that merely capturing the static importance of individual experts is insufficient to identify and preserve the complex expert combination paths required for some specific tasks.
Ultimately, the combination of both metrics achieves the highest average accuracy, confirming that TI and IS provide complementary information essential for maintaining model capabilities across diverse domains.

\begin{figure*}[t]
\centering
\includegraphics[width=1\linewidth, trim=0 10 0 0]{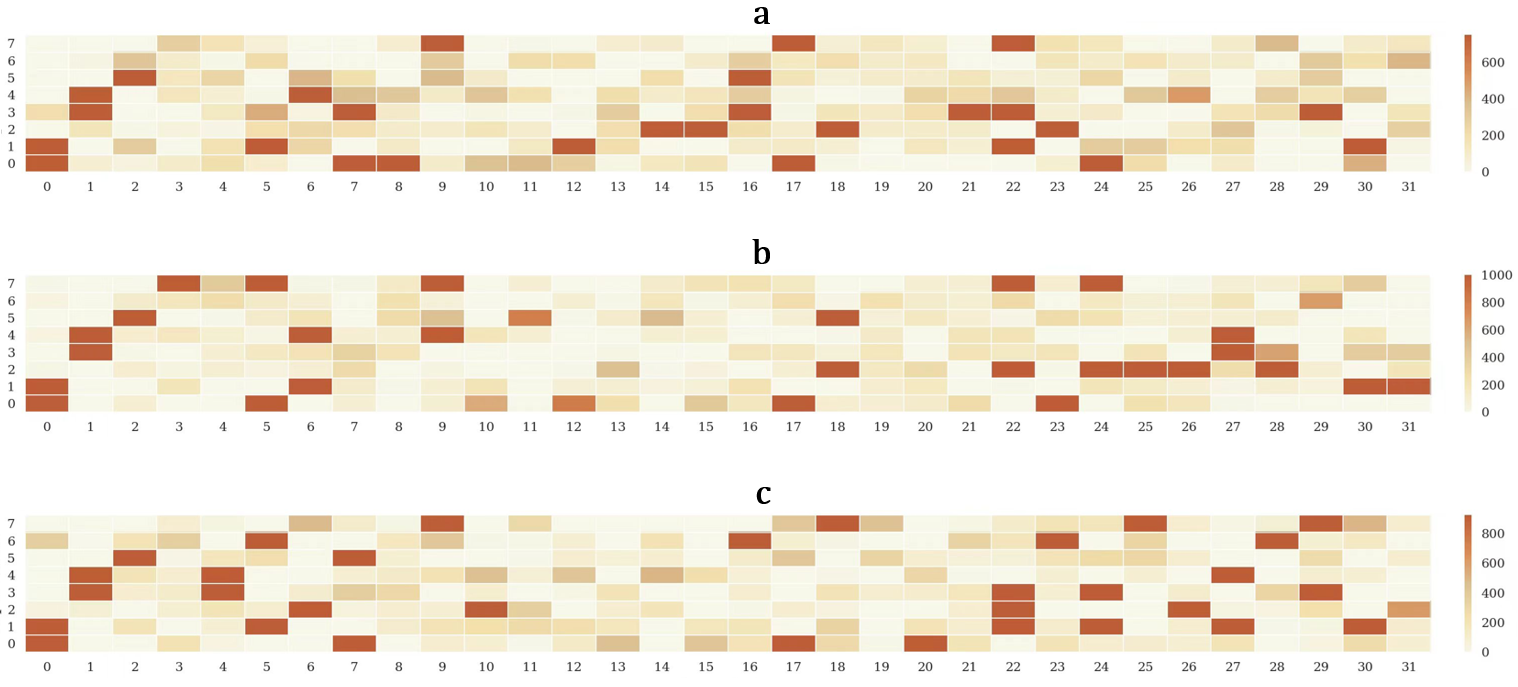}
\caption{
Visualization of expert importance for the Mixtral-8x7B model.
These heatmaps illustrate the layer-wise and expert-specific importance, quantified by the frequency an expert is selected during the path planning phase. Calibration data are partitioned into $K=10$ clusters, and expert frequencies are computed by sampling the top-100 paths for each cluster. An expert's importance is determined by its selection frequency, where a higher selection frequency (indicated by a darker color) signifies greater importance. The results include a general-domain task, MMLU (top), and two domain-specific tasks, MedQA (middle) and GSM8K (bottom).
}
\label{fig:importance}
\vspace{-0.3cm}
\end{figure*}

\subsection{Effect of the Number of Clusters}

\begin{figure}[t]
\centering
\includegraphics[width=1\linewidth, trim=0 15 0 0]{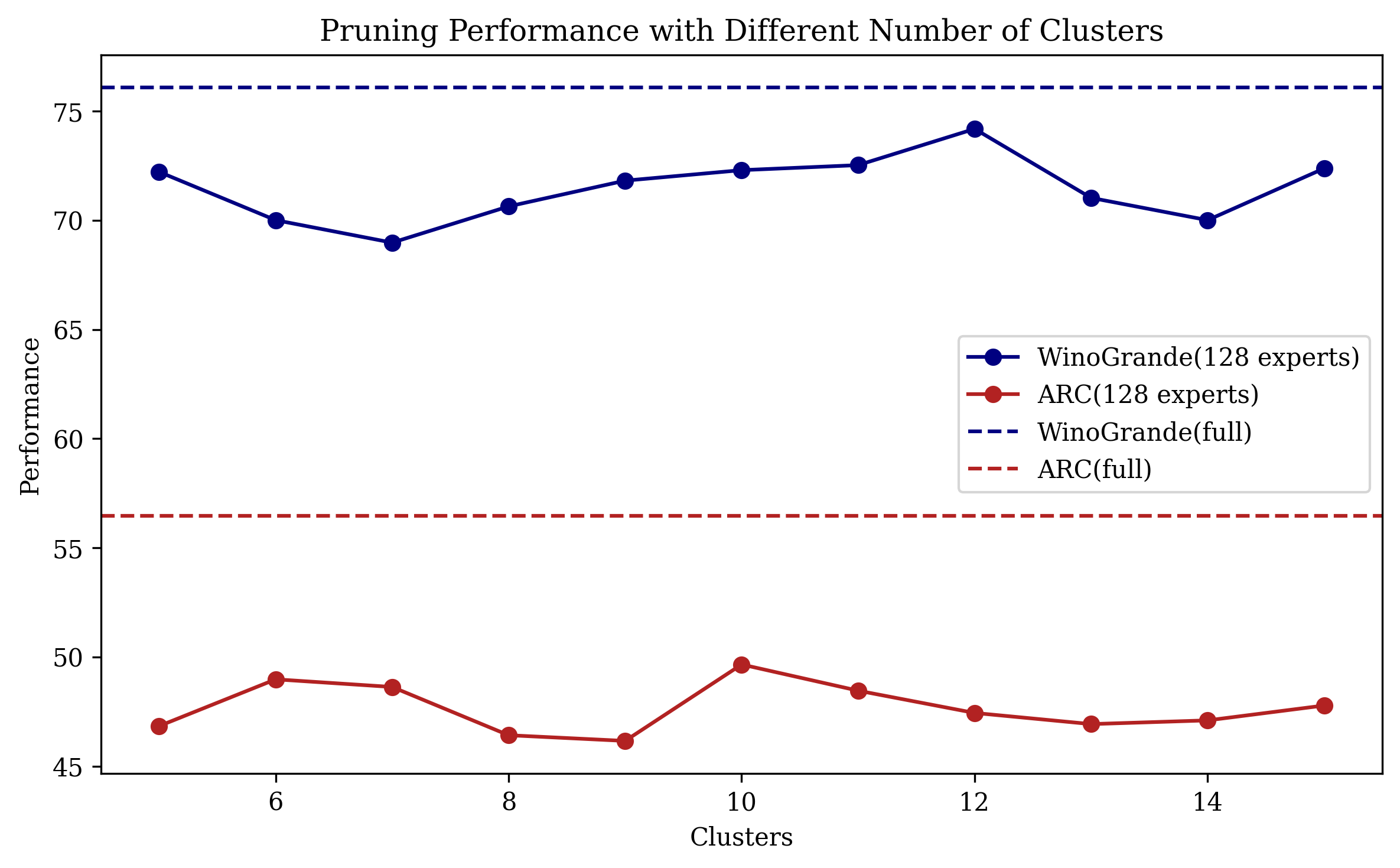}
\caption{
\textcolor{black}{
Effect of number of clusters ($K$) on the performance of the Mixtral-8x7B model pruned to 50\% expert sparsity.
The horizontal axis denotes the number of k-means clusters ($K$) used to construct calibration data, and the vertical axis shows the model's accuracy on WinoGrande (blue) and ARC (red).}
}
\label{fig:number}
\vspace{-0.4cm}
\end{figure}

As described in section \ref{sec:dataset}, our approach for constructing calibration data involves partitioning the dataset using k-means clustering and selecting representative samples.
In this section, we investigate the impact of the number of clusters ($K$) on the pruned model's performance of the Mixtral-8x7B architecture.
In particular, we prune the model to 50\% expert sparsity using calibration data constructed with varying numbers of clusters, and visualize the resulting pruning performance.
The performance variations on WinoGrande and ARC are presented in Figure \ref{fig:number}.
The horizontal axis represents the number of k-means clusters ($K$) used to construct calibration data, and the vertical axis shows the pruned model's performance measured by accuracy.
Several trends are observed.
First, across both datasets, the pruned model exhibits a unimodal sensitivity to the choice of $K$, with performance increasing up to a critical point and deteriorating thereafter.  
This trend highlights the critical trade-off between the diversity and representativeness of the calibration data.
When $K$ is too small, the clustering is overly coarse, and the selected representative samples fail to capture the full diversity of the entire dataset, resulting in incomplete calibration gradient information.
Conversely, when $K$ is too large, the clustering becomes overly granular, potentially introducing noise or information redundancy, thereby diminishing the efficacy of the calibration information.
Second, the optimal number of clusters differs between the two tasks ($K$=12 for WinoGrande and $K$=10 for ARC).
This discrepancy reflects the inherent structural complexity and feature space heterogeneity of the different datasets, suggesting that building a high-quality calibration set is task-dependent.

\subsection{Visualization of Expert Importance}

To better understand expert specialization within MoE architectures, we quantify and visualize expert importance based on their selection frequency during the path planning phase.
Specifically, we conduct experiments using the Mixtral-8x7B model. 
We perform clustering ($K$=10) on calibration data derived from three distinct benchmarks: MMLU, GSM8K, and MedQA.
During the routing process, we record the selection frequency of each expert by sampling the top-$m$ paths (where $m$=100).
Experts identified as outliers in the data pre-processing stage were assigned the maximum selection count to reflect their highest potential importance.
The aggregated results for each task are presented as heatmaps; in these visualizations, darker colors correlate with higher selection frequency, thereby indicating the relative importance of an expert for a given task.
The results are presented in Figure \ref{fig:importance}.
Two key findings are observed.
First, the expert importance profiles are consistent across MMLU, MedQA, and GSM8K in the shallow layers (i.e., first to third layers).
Conversely, a significant divergence is observed in the deep layers (i.e., 25th to 31st layers).
Specific domain tasks (MedQA and GSM8K) demonstrate that critical importance is concentrated within a small subset of experts (e.g., MedQA's highest importance relies on the first and second experts in the deep blocks).
In contrast, the general domain task MMLU maintains a more distributed importance pattern throughout the deeper layers.
This indicates that the initial processing stages utilize a shared, domain-agnostic set of experts, and that the model achieves expert hyper-specialization only in the deeper layers.
In these deep blocks, a few experts become disproportionately crucial for task-specific computation beyond foundational feature extraction.
Second, at the 23rd layer, the first expert is crucial for the MedQA task (high selection frequency), but its contribution is negligible for GSM8K (selection frequency near zero). 
Similar phenomena are observed in several layers, such as the 18th, 25th, and 26th.
This stark difference in the importance of the same expert across specific layers strongly suggests the orthogonality or task-isolation of expert functions within the MoE architecture.

\section{Related Work}

LLMs have achieved remarkable performance on many tasks in artificial intelligence \cite{brown2020language,li2020conditional,diao-etal-2020-zen,song-etal-2020-summarizing,chen-etal-2021-relation,diao2021tilgan,qin-2022-enhancing,touvron2023llama,gan2023ziya2,liu2024bootstrapping,li2024challenging,yang2025qwen3,su2025text}.
The MoE architecture has become a mainstream paradigm in LLMs, enabling exponential parameter scaling through sparse activation \cite{tian2024dialogue,guo2025deepseek,cai2025survey}.
However, MoE models still require loading all expert weights during inference, resulting in a substantial memory footprint that makes compression indispensable for practical deployment.
Current MoE compression strategies fall into three major categories: dynamic expert selection, expert merging, and expert pruning.
Dynamic expert selection introduces zero-computation experts or estimates token-level importance to dynamically adjust the number of activated experts during inference \cite{jin2024moe++,raposo2024mixture,zeng2024adamoe,yue2024ada,team2025longcat}.
However, since the total parameter count remains unchanged, these approaches do not fundamentally reduce memory usage, limiting their utility as true compression approaches.
Expert merging reduces the number of experts by linearly combining weights, but the criteria it relies on—such as weight similarity—are often oversimplified and brittle \cite{li2023merge,he2023merging,chen2024retraining,li2025sub}.
As a result, merged experts frequently fail to preserve the specialized knowledge of their originals, causing unpredictable performance degradation.
For expert pruning approaches, they differ from conventional pruning models \cite{frantar2023sparsegpt,sun2023simple,tian2025frustratingly} since the expert pruning studies remove parameters at the expert-level, where all parameters in an expert are pruned if the expert is determined to be pruned.
These expert pruning studies initially rely on perturbation-based analysis, measuring output differences before and after removing each expert \cite{cao2024condense,lu2024not}.
Later studies propose more efficient metrics such as routing frequency and activation norms for expert selection \cite{muzio2024seer,chowdhury2024provably,dong2025domain,lasby2025reap,zhou2025dropping}.
These developments have substantially improved pruning efficiency and performance, yet most metrics are inherently local and fail to capture an expert’s true global contribution to task performance.
Moreover, most prior pruning strategies adopt a uniform pruning ratio across layers, overlooking the heterogeneous contribution of experts in different layers.
Different from existing studies, we propose a pruning approach based on global optimal trajectory planning.
Our approach integrates perturbation signals, routing decisions, and expert activations into a unified global graph to evaluate expert importance.
By optimizing expert selection over cross-layer computational paths, the framework adaptively determines the number of experts to retain in each layer, enabling efficient and naturally non-uniform pruning.

\section{Conclusion}

In this paper, we presented a novel trajectory-driven expert pruning approach for domain-specific MoE model compression.
By reformulating the MoE architecture as a directed weighted graph, our approach provides a global perspective on cross-layer expert dependencies.
To achieve fine-grained expert evaluation, we introduced a multi-granularity weighting scheme that integrates perturbation-based loss, routing probabilities, and expert activations.
Building on this representation, we employed a dynamic-programming-based global path planning algorithm to identify the critical computational paths and retain only the experts that lie along them.
Experiment results demonstrate that our approach outperforms existing pruning baselines across multiple benchmarks, preserving both general-domain and domain-specific capabilities.
Overall, this work offers a unified, efficient, and interpretable solution for task-aware MoE compression.

\nocite{langley00}

\bibliography{example_paper}

@article{jiang2024mixtral,
  title={Mixtral of experts},
  author={Jiang, Albert Q and Sablayrolles, Alexandre and Roux, Antoine and Mensch, Arthur and Savary, Blanche and Bamford, Chris and Chaplot, Devendra Singh and Casas, Diego de las and Hanna, Emma Bou and Bressand, Florian and others},
  journal={arXiv preprint arXiv:2401.04088},
  year={2024}
}

@inproceedings{frantar2023sparsegpt,
  title={Sparsegpt: Massive language models can be accurately pruned in one-shot},
  author={Frantar, Elias and Alistarh, Dan},
  booktitle={International conference on machine learning},
  pages={10323--10337},
  year={2023},
  organization={PMLR}
}

@article{gan2023ziya2,
  title={Ziya2: Data-centric learning is all llms need},
  author={Gan, Ruyi and Wu, Ziwei and Sun, Renliang and Lu, Junyu and Wu, Xiaojun and Zhang, Dixiang and Pan, Kunhao and He, Junqing and Tian, Yuanhe and Yang, Ping and others},
  journal={arXiv preprint arXiv:2311.03301},
  year={2023}
}

@article{li2020conditional,
  title={Conditional augmentation for aspect term extraction via masked sequence-to-sequence generation},
  author={Li, Kun and Chen, Chengbo and Quan, Xiaojun and Ling, Qing and Song, Yan},
  journal={arXiv preprint arXiv:2004.14769},
  year={2020}
}

@inproceedings{diao2021tilgan,
  title={TILGAN: transformer-based implicit latent GAN for diverse and coherent text generation},
  author={Diao, Shizhe and Shen, Xinwei and Shum, Kashun and Song, Yan and Zhang, Tong},
  booktitle={Findings of the Association for Computational linguistics: ACL-IJCNLP 2021},
  pages={4844--4858},
  year={2021}
}

@inproceedings{liu2024bootstrapping,
  title={Bootstrapping large language models for radiology report generation},
  author={Liu, Chang and Tian, Yuanhe and Chen, Weidong and Song, Yan and Zhang, Yongdong},
  booktitle={Proceedings of the AAAI Conference on Artificial Intelligence},
  volume={38},
  number={17},
  pages={18635--18643},
  year={2024}
}

@article{hendrycks2020measuring,
  title={Measuring massive multitask language understanding},
  author={Hendrycks, Dan and Burns, Collin and Basart, Steven and Zou, Andy and Mazeika, Mantas and Song, Dawn and Steinhardt, Jacob},
  journal={arXiv preprint arXiv:2009.03300},
  year={2020}
}

@article{jin2021disease,
  title={What disease does this patient have? a large-scale open domain question answering dataset from medical exams},
  author={Jin, Di and Pan, Eileen and Oufattole, Nassim and Weng, Wei-Hung and Fang, Hanyi and Szolovits, Peter},
  journal={Applied Sciences},
  volume={11},
  number={14},
  pages={6421},
  year={2021},
  publisher={MDPI}
}

@article{clark2018think,
  title={Think you have solved question answering? try arc, the ai2 reasoning challenge},
  author={Clark, Peter and Cowhey, Isaac and Etzioni, Oren and Khot, Tushar and Sabharwal, Ashish and Schoenick, Carissa and Tafjord, Oyvind},
  journal={arXiv preprint arXiv:1803.05457},
  year={2018}
}

@article{sun2023simple,
  title={A simple and effective pruning approach for large language models},
  author={Sun, Mingjie and Liu, Zhuang and Bair, Anna and Kolter, J Zico},
  journal={arXiv preprint arXiv:2306.11695},
  year={2023}
}

@article{brown2020language,
  title={Language {M}odels are {F}ew-shot {L}earners},
  author={Brown, Tom and Mann, Benjamin and Ryder, Nick and Subbiah, Melanie and Kaplan, Jared D and Dhariwal, Prafulla and Neelakantan, Arvind and Shyam, Pranav and Sastry, Girish and Askell, Amanda and others},
  journal={Advances in neural information processing systems},
  volume={33},
  pages={1877--1901},
  year={2020}
}

@article{artetxe2021efficient,
  title={Efficient large scale language modeling with mixtures of experts},
  author={Artetxe, Mikel and Bhosale, Shruti and Goyal, Naman and Mihaylov, Todor and Ott, Myle and Shleifer, Sam and Lin, Xi Victoria and Du, Jingfei and Iyer, Srinivasan and Pasunuru, Ramakanth and others},
  journal={arXiv preprint arXiv:2112.10684},
  year={2021}
}

@article{muzio2024seer,
  title={Seer-moe: Sparse expert efficiency through regularization for mixture-of-experts},
  author={Muzio, Alexandre and Sun, Alex and He, Churan},
  journal={arXiv preprint arXiv:2404.05089},
  year={2024}
}

@article{cao2024condense,
  title={Condense, Don't Just Prune: Enhancing Efficiency and Performance in MoE Layer Pruning},
  author={Cao, Mingyu and Li, Gen and Ji, Jie and Zhang, Jiaqi and Ma, Xiaolong and Liu, Shiwei and Yin, Lu},
  journal={arXiv preprint arXiv:2412.00069},
  year={2024}
}

@article{lu2024not,
  title={Not all experts are equal: Efficient expert pruning and skipping for mixture-of-experts large language models},
  author={Lu, Xudong and Liu, Qi and Xu, Yuhui and Zhou, Aojun and Huang, Siyuan and Zhang, Bo and Yan, Junchi and Li, Hongsheng},
  journal={arXiv preprint arXiv:2402.14800},
  year={2024}
}

@article{dong2025domain,
  title={Domain-Specific Pruning of Large Mixture-of-Experts Models with Few-shot Demonstrations},
  author={Dong, Zican and Peng, Han and Liu, Peiyu and Zhao, Wayne Xin and Wu, Dong and Xiao, Feng and Wang, Zhifeng},
  journal={arXiv preprint arXiv:2504.06792},
  year={2025}
}

@article{fedus2022switch,
  title={Switch transformers: Scaling to trillion parameter models with simple and efficient sparsity},
  author={Fedus, William and Zoph, Barret and Shazeer, Noam},
  journal={Journal of Machine Learning Research},
  volume={23},
  number={120},
  pages={1--39},
  year={2022}
}

@article{dai2024deepseekmoe,
  title={Deepseekmoe: Towards ultimate expert specialization in mixture-of-experts language models},
  author={Dai, Damai and Deng, Chengqi and Zhao, Chenggang and Xu, RX and Gao, Huazuo and Chen, Deli and Li, Jiashi and Zeng, Wangding and Yu, Xingkai and Wu, Yu and others},
  journal={arXiv preprint arXiv:2401.06066},
  year={2024}
}

@article{achiam2023gpt,
  title={{GPT-4 technical report}},
  author={Achiam, Josh and Adler, Steven and Agarwal, Sandhini and Ahmad, Lama and Akkaya, Ilge and Aleman, Florencia Leoni and Almeida, Diogo and Altenschmidt, Janko and Altman, Sam and Anadkat, Shyamal and others},
  journal={arXiv preprint arXiv:2303.08774},
  year={2023}
}

@article{cobbe2021gsm8k,
  title={Training Verifiers to Solve Math Word Problems},
  author={Cobbe, Karl and Kosaraju, Vineet and Bavarian, Mohammad and Chen, Mark and Jun, Heewoo and Kaiser, Lukasz and Plappert, Matthias and Tworek, Jerry and Hilton, Jacob and Nakano, Reiichiro and Hesse, Christopher and Schulman, John},
  journal={arXiv preprint arXiv:2110.14168},
  year={2021}
}

@article{guo2025deepseek,
  title={Deepseek-r1: Incentivizing reasoning capability in llms via reinforcement learning},
  author={Guo, Daya and Yang, Dejian and Zhang, Haowei and Song, Junxiao and Zhang, Ruoyu and Xu, Runxin and Zhu, Qihao and Ma, Shirong and Wang, Peiyi and Bi, Xiao and others},
  journal={arXiv preprint arXiv:2501.12948},
  year={2025}
}

@article{cai2025survey,
  title={A survey on mixture of experts in large language models},
  author={Cai, Weilin and Jiang, Juyong and Wang, Fan and Tang, Jing and Kim, Sunghun and Huang, Jiayi},
  journal={IEEE Transactions on Knowledge and Data Engineering},
  year={2025},
  publisher={IEEE}
}

@article{li2023merge,
  title={Merge, then compress: Demystify efficient smoe with hints from its routing policy},
  author={Li, Pingzhi and Zhang, Zhenyu and Yadav, Prateek and Sung, Yi-Lin and Cheng, Yu and Bansal, Mohit and Chen, Tianlong},
  journal={arXiv preprint arXiv:2310.01334},
  year={2023}
}

@article{he2023merging,
  title={Merging experts into one: Improving computational efficiency of mixture of experts},
  author={He, Shwai and Fan, Run-Ze and Ding, Liang and Shen, Li and Zhou, Tianyi and Tao, Dacheng},
  journal={arXiv preprint arXiv:2310.09832},
  year={2023}
}

@article{jin2024moe++,
  title={Moe++: Accelerating mixture-of-experts methods with zero-computation experts},
  author={Jin, Peng and Zhu, Bo and Yuan, Li and Yan, Shuicheng},
  journal={arXiv preprint arXiv:2410.07348},
  year={2024}
}

@article{raposo2024mixture,
  title={Mixture-of-depths: Dynamically allocating compute in transformer-based language models},
  author={Raposo, David and Ritter, Sam and Richards, Blake and Lillicrap, Timothy and Humphreys, Peter Conway and Santoro, Adam},
  journal={arXiv preprint arXiv:2404.02258},
  year={2024}
}

@article{zeng2024adamoe,
  title={Adamoe: Token-adaptive routing with null experts for mixture-of-experts language models},
  author={Zeng, Zihao and Miao, Yibo and Gao, Hongcheng and Zhang, Hao and Deng, Zhijie},
  journal={arXiv preprint arXiv:2406.13233},
  year={2024}
}

@inproceedings{yue2024ada,
  title={Ada-k routing: Boosting the efficiency of moe-based llms},
  author={Yue, Tongtian and Guo, Longteng and Cheng, Jie and Gao, Xuange and Huang, Hua and Liu, Jing},
  booktitle={The Thirteenth International Conference on Learning Representations},
  year={2024}
}

@article{team2025longcat,
  title={Longcat-flash technical report},
  author={Team, Meituan LongCat and Li, Bei and Lei, Bingye and Wang, Bo and Rong, Bolin and Wang, Chao and Zhang, Chao and Gao, Chen and Zhang, Chen and Sun, Cheng and others},
  journal={arXiv preprint arXiv:2509.01322},
  year={2025}
}

@article{guo2025cluster,
  title={Cluster-Driven Expert Pruning for Mixture-of-Experts Large Language Models},
  author={Guo, Hongcheng and Yao, Juntao and Wang, Boyang and Du, Junjia and Cao, Shaosheng and Di, Donglin and Zhang, Shun and Li, Zhoujun},
  journal={arXiv preprint arXiv:2504.07807},
  year={2025}
}

@article{zhang2025mone,
  title={MoNE: Replacing Redundant Experts with Lightweight Novices for Structured Pruning of MoE},
  author={Zhang, Geng and Han, Yuxuan and Lou, Yuxuan and Zhao, Wangbo and Zhang, Yiqi and You, Yang},
  journal={arXiv preprint arXiv:2507.00390},
  year={2025}
}

@article{tian2025frustratingly,
  title={Frustratingly Easy Task-aware Pruning for Large Language Models},
  author={Tian, Yuanhe and Liu, Junjie and Yang, Xican and Ye, Haishan and Song, Yan},
  journal={arXiv preprint arXiv:2510.22489},
  year={2025}
}

@article{su2025text,
  title={Text Reinforcement for Multimodal Time Series Forecasting},
  author={Su, Chen and Tian, Yuanhe and Song, Yan and Zhang, Yongdong},
  journal={arXiv preprint arXiv:2509.00687},
  year={2025}
}

@inproceedings{li2024challenging,
  title={Challenging large language models with new tasks: A study on their adaptability and robustness},
  author={Li, Chenxi and Tian, Yuanhe and Zerong, Zhaxi and Song, Yan and Xia, Fei},
  booktitle={Findings of the Association for Computational Linguistics ACL 2024},
  pages={8140--8162},
  year={2024}
}

@inproceedings{diao-etal-2020-zen,
    title = "{ZEN}: {P}re-training {C}hinese {T}ext {E}ncoder {E}nhanced by {N}-gram {R}epresentations",
    author = "Diao, Shizhe  and
      Bai, Jiaxin  and
      Song, Yan  and
      Zhang, Tong  and
      Wang, Yonggang",
    booktitle = "Findings of the Association for Computational Linguistics: EMNLP 2020",
    month = nov,
    year = "2020",
    pages = "4729--4740",
}

@inproceedings{song-etal-2020-summarizing,
    title = "Summarizing {M}edical {C}onversations via {I}dentifying {I}mportant {U}tterances",
    author = "Song, Yan  and
      Tian, Yuanhe  and
      Wang, Nan  and
      Xia, Fei",
    booktitle = "Proceedings of the 28th International Conference on Computational Linguistics",
    month = dec,
    year = "2020",
    pages = "717--729",
}

@inproceedings{chen-etal-2021-relation,
    title = "Relation {E}xtraction with {T}ype-aware {M}ap {M}emories of {W}ord {D}ependencies",
    author = "Chen, Guimin  and
      Tian, Yuanhe  and
      Song, Yan  and
      Wan, Xiang",
    booktitle = "Findings of the Association for Computational Linguistics: ACL-IJCNLP 2021",
    month = aug,
    year = "2021",
    address = "Online",
    pages = "2501--2512",
}

@inproceedings{tian2024dialogue,
  title={Dialogue summarization with mixture of experts based on large language models},
  author={Tian, Yuanhe and Xia, Fei and Song, Yan},
  booktitle={Proceedings of the 62nd Annual Meeting of the Association for Computational Linguistics (Volume 1: Long Papers)},
  pages={7143--7155},
  year={2024}
}

@inproceedings{qin-2022-enhancing,
    title = "Enhancing {R}elation {E}xtraction via {A}dversarial {M}ulti-task {L}earning",
    author = "Qin, Han and
    Tian, Yuanhe  and
      Song, Yan",
    booktitle = "Proceedings of the 13th Language Resources and Evaluation Conference",
    month = June,
    year = "2022",
}

@article{zellers2019hellaswag,
  title={Hellaswag: Can a machine really finish your sentence?},
  author={Zellers, Rowan and Holtzman, Ari and Bisk, Yonatan and Farhadi, Ali and Choi, Yejin},
  journal={arXiv preprint arXiv:1905.07830},
  year={2019}
}

@article{sakaguchi2021winogrande,
  title={Winogrande: An adversarial winograd schema challenge at scale},
  author={Sakaguchi, Keisuke and Bras, Ronan Le and Bhagavatula, Chandra and Choi, Yejin},
  journal={Communications of the ACM},
  volume={64},
  number={9},
  pages={99--106},
  year={2021},
  publisher={ACM New York, NY, USA}
}

@article{su2025unveiling,
  title={Unveiling super experts in mixture-of-experts large language models},
  author={Su, Zunhai and Li, Qingyuan and Zhang, Hao and Qian, YuLei and Xie, Yuchen and Yuan, Kehong},
  journal={arXiv preprint arXiv:2507.23279},
  year={2025}
}

@article{chen2024retraining,
  title={Retraining-Free Merging of Sparse MoE via Hierarchical Clustering},
  author={Chen, I and Liu, Hsu-Shen and Sun, Wei-Fang and Chao, Chen-Hao and Hsu, Yen-Chang and Lee, Chun-Yi and others},
  journal={arXiv preprint arXiv:2410.08589},
  year={2024}
}

@article{he2024demystifying,
  title={Demystifying the compression of mixture-of-experts through a unified framework},
  author={He, Shwai and Dong, Daize and Ding, Liang and Li, Ang},
  journal={arXiv e-prints},
  pages={arXiv--2406},
  year={2024}
}

@inproceedings{lee2025stun,
  title={Stun: Structured-then-unstructured pruning for scalable moe pruning},
  author={Lee, Jaeseong and Hwang, Seung-won and Qiao, Aurick and Campos, Daniel F and Yao, Zhewei and He, Yuxiong},
  booktitle={Proceedings of the 63rd Annual Meeting of the Association for Computational Linguistics (Volume 1: Long Papers)},
  pages={13660--13676},
  year={2025}
}

@article{jaiswal2025finding,
  title={Finding Fantastic Experts in MoEs: A Unified Study for Expert Dropping Strategies and Observations},
  author={Jaiswal, Ajay and Wang, Jianyu and Li, Yixiao and Li, Pingzhi and Chen, Tianlong and Wang, Zhangyang and Wang, Chong and Pang, Ruoming and Du, Xianzhi},
  journal={arXiv preprint arXiv:2504.05586},
  year={2025}
}

@article{zhou2025dropping,
  title={Dropping Experts, Recombining Neurons: Retraining-Free Pruning for Sparse Mixture-of-Experts LLMs},
  author={Zhou, Yixiao and Zhao, Ziyu and Cheng, Dongzhou and Gui, Jie and Yang, Yi and Wu, Fei and Cheng, Yu and Fan, Hehe and others},
  journal={arXiv preprint arXiv:2509.10377},
  year={2025}
}

@article{li2025sub,
  title={Sub-MoE: Efficient Mixture-of-Expert LLMs Compression via Subspace Expert Merging},
  author={Li, Lujun and Qiyuan, Zhu and Wang, Jiacheng and Li, Wei and Gu, Hao and Han, Sirui and Guo, Yike},
  journal={arXiv preprint arXiv:2506.23266},
  year={2025}
}

@article{lasby2025reap,
  title={REAP the Experts: Why Pruning Prevails for One-Shot MoE compression},
  author={Lasby, Mike and Lazarevich, Ivan and Sinnadurai, Nish and Lie, Sean and Ioannou, Yani and Thangarasa, Vithursan},
  journal={arXiv preprint arXiv:2510.13999},
  year={2025}
}

@article{touvron2023llama,
  title={Llama 2: Open foundation and fine-tuned chat models},
  author={Touvron, Hugo and Martin, Louis and Stone, Kevin and Albert, Peter and Almahairi, Amjad and Babaei, Yasmine and Bashlykov, Nikolay and Batra, Soumya and Bhargava, Prajjwal and Bhosale, Shruti and others},
  journal={arXiv preprint arXiv:2307.09288},
  year={2023}
}

@article{wei2022emergent,
  title={Emergent abilities of large language models},
  author={Wei, Jason and Tay, Yi and Bommasani, Rishi and Raffel, Colin and Zoph, Barret and Borgeaud, Sebastian and Yogatama, Dani and Bosma, Maarten and Zhou, Denny and Metzler, Donald and others},
  journal={arXiv preprint arXiv:2206.07682},
  year={2022}
}

@article{chowdhury2024provably,
  title={A provably effective method for pruning experts in fine-tuned sparse mixture-of-experts},
  author={Chowdhury, Mohammed Nowaz Rabbani and Wang, Meng and Maghraoui, Kaoutar El and Wang, Naigang and Chen, Pin-Yu and Carothers, Christopher},
  journal={arXiv preprint arXiv:2405.16646},
  year={2024}
}

@article{yang2025qwen3,
  title={Qwen3 technical report},
  author={Yang, An and Li, Anfeng and Yang, Baosong and Zhang, Beichen and Hui, Binyuan and Zheng, Bo and Yu, Bowen and Gao, Chang and Huang, Chengen and Lv, Chenxu and others},
  journal={arXiv preprint arXiv:2505.09388},
  year={2025}
}

@article{meta2025llama,
  title={The llama 4 herd: The beginning of a new era of natively multimodal ai innovation},
  author={Meta, AI},
  journal={https://ai. meta. com/blog/llama-4-multimodal-intelligence/, checked on},
  volume={4},
  number={7},
  pages={2025},
  year={2025}
}

@article{liu2024deepseek,
  title={Deepseek-v3 technical report},
  author={Liu, Aixin and Feng, Bei and Xue, Bing and Wang, Bingxuan and Wu, Bochao and Lu, Chengda and Zhao, Chenggang and Deng, Chengqi and Zhang, Chenyu and Ruan, Chong and others},
  journal={arXiv preprint arXiv:2412.19437},
  year={2024}
}

@article{team2025kimi,
  title={Kimi k2: Open agentic intelligence},
  author={Team, Kimi and Bai, Yifan and Bao, Yiping and Chen, Guanduo and Chen, Jiahao and Chen, Ningxin and Chen, Ruijue and Chen, Yanru and Chen, Yuankun and Chen, Yutian and others},
  journal={arXiv preprint arXiv:2507.20534},
  year={2025}
}

@article{lepikhin2020gshard,
  title={Gshard: Scaling giant models with conditional computation and automatic sharding},
  author={Lepikhin, Dmitry and Lee, HyoukJoong and Xu, Yuanzhong and Chen, Dehao and Firat, Orhan and Huang, Yanping and Krikun, Maxim and Shazeer, Noam and Chen, Zhifeng},
  journal={arXiv preprint arXiv:2006.16668},
  year={2020}
}

@article{liu2023visual,
  title={Visual instruction tuning},
  author={Liu, Haotian and Li, Chunyuan and Wu, Qingyang and Lee, Yong Jae},
  journal={Advances in neural information processing systems},
  volume={36},
  pages={34892--34916},
  year={2023}
}

@article{li2023starcoder,
  title={Starcoder: may the source be with you!},
  author={Li, Raymond and Allal, Loubna Ben and Zi, Yangtian and Muennighoff, Niklas and Kocetkov, Denis and Mou, Chenghao and Marone, Marc and Akiki, Christopher and Li, Jia and Chim, Jenny and others},
  journal={arXiv preprint arXiv:2305.06161},
  year={2023}
}

@article{lin2023evolutionary,
  title={Evolutionary-scale prediction of atomic-level protein structure with a language model},
  author={Lin, Zeming and Akin, Halil and Rao, Roshan and Hie, Brian and Zhu, Zhongkai and Lu, Wenting and Smetanin, Nikita and Verkuil, Robert and Kabeli, Ori and Shmueli, Yaniv and others},
  journal={Science},
  volume={379},
  number={6637},
  pages={1123--1130},
  year={2023},
  publisher={American Association for the Advancement of Science}
}

@inproceedings{lo2025closer,
  title={A closer look into mixture-of-experts in large language models},
  author={Lo, Ka Man and Huang, Zeyu and Qiu, Zihan and Wang, Zili and Fu, Jie},
  booktitle={Findings of the Association for Computational Linguistics: NAACL 2025},
  pages={4427--4447},
  year={2025}
}

@article{zeng2025glm,
  title={Glm-4.5: Agentic, reasoning, and coding (arc) foundation models},
  author={Zeng, Aohan and Lv, Xin and Zheng, Qinkai and Hou, Zhenyu and Chen, Bin and Xie, Chengxing and Wang, Cunxiang and Yin, Da and Zeng, Hao and Zhang, Jiajie and others},
  journal={arXiv preprint arXiv:2508.06471},
  year={2025}
}
\bibliographystyle{icml2026}

\newpage
\appendix
\onecolumn

\end{document}